\documentclass[10pt,twocolumn,letterpaper]{article}

\usepackage{cvpr}              
\definecolor{cvprblue}{rgb}{0.21,0.49,0.74}
\usepackage[pagebackref,breaklinks,colorlinks,allcolors=cvprblue]{hyperref}
\usepackage{multirow}
\usepackage{graphicx}
\usepackage[accsupp]{axessibility}

\begin{document}

\def\paperID{122} 
\def\confName{CVPR}
\def\confYear{2025}

\title{Towards Holistic Visual Quality Assessment of AI-Generated Videos: A LLM-Based Multi-Dimensional Evaluation Model}

\author{Zelu Qi, Ping Shi\thanks{Corresponding author}, Chaoyang Zhang, Shuqi Wang, Fei Zhao, Da Pan, Zefeng Ying \\
Communication University of China, No. 1 Dingfuzhuang East Street, Beijing, China\\
{\tt\small theoneqi2001@cuc.edu.cn, shiping@cuc.edu.cn}
}
\maketitle

\begin{abstract}

The development of AI-Generated Video (AIGV) technology has been remarkable in recent years, significantly transforming the paradigm of video content production. However, AIGVs still suffer from noticeable visual quality defects, such as noise, blurriness, frame jitter and low dynamic degree, which severely impact the user's viewing experience. Therefore, an effective automatic visual quality assessment is of great importance for AIGV content regulation and generative model improvement. In this work, we decompose the visual quality of AIGVs into three dimensions: technical quality, motion quality, and video semantics. For each dimension, we design corresponding encoder to achieve effective feature representation. Moreover, considering the outstanding performance of large language models (LLMs) in various vision and language tasks, we introduce a LLM as the quality regression module. To better enable the LLM to establish reasoning associations between multi-dimensional features and visual quality, we propose a specially designed multi-modal prompt engineering framework. Additionally, we incorporate LoRA fine-tuning technology during the training phase, allowing the LLM to better adapt to specific tasks. Our proposed method achieved \textbf{second place} in the NTIRE 2025 Quality Assessment of AI-Generated Content Challenge: Track 2 AI Generated video, demonstrating its effectiveness. Codes can be obtained at \href{https://github.com/QiZelu/AIGVEval}{AIGVEval}.

\end{abstract}

\section{Introduction}
\label{sec:intro}

\begin{figure}[htbp]
    \centering
    \includegraphics[width=1.0\linewidth]{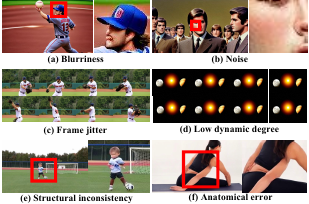}
    \caption{Typical distortion types of AI-Generated videos. (a) blurriness, (b) noise, (c) frame jitter, (d) low dynamic degree, (e) structural inconsistency and (f) anatomical error.}
    \label{fig:fig1}
\end{figure}

In recent years, generative artificial intelligence (GenAI), represented by text-to-video (T2V) technology, has garnered significant attention, leading to the emergence of numerous advanced T2V models. These T2V models enable users to generate videos with simple and free-form textual prompts, greatly simplifying the video production process. However, constrained by current technological limitations, AI-Generated videos (AIGVs) still exhibit noticeable quality defects. Particularly in terms of visual quality, as illustrated in Figure \ref{fig:fig1}, AIGVs often suffer from significant spatio-temporal defects, such as blurriness or low dynamic degree, which severely degrade the user's viewing experience. Therefore, accurately assessing the visual quality of AIGVs is essential for ensuring an optimal user experience and guiding the optimization of T2V models.

\begin{figure*}[bhtp]
    \centering
    \includegraphics[width=0.8\linewidth]{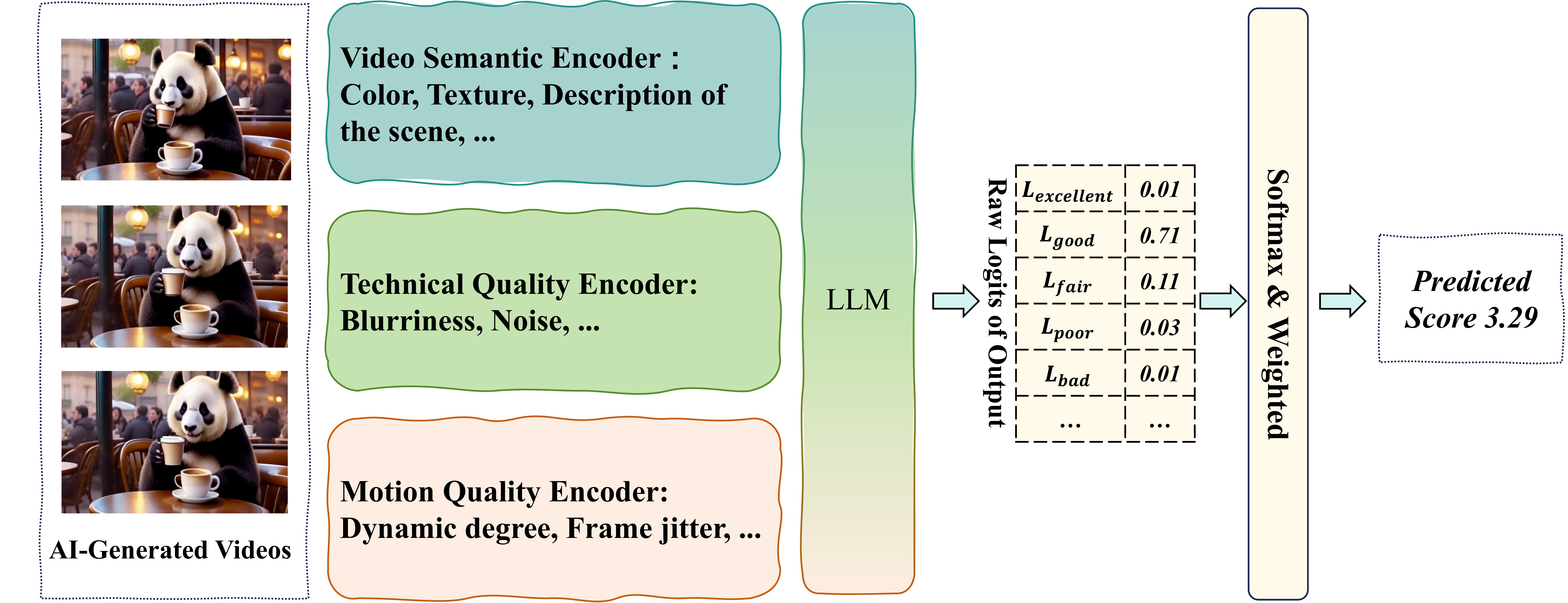}
    \caption{The framework of our method.}
    \label{fig:fig2}
\end{figure*}

With the rapid development of large language models (LLMs), they exhibit strong prior knowledge and exceptional adaptability across various vision and language tasks. Consequently, numerous studies have emerged that leverage LLMs for video quality assessment. For instance, T2VQA\cite{kou2024subjective} and Q-Eval-Score\cite{zhang2025q} have applied LLMs to AIGV quality evaluation, achieving competitive performance.

However, we argue that these approaches don't fully exploit the potential of LLMs in quality assessment tasks. This limitation primarily lies in two aspects: \textbf{1) The pre-encoders fail to effectively capture motion information within videos. 2) The prompts cannot effectively help LLMs understand the specific meaning of the features extracted by the encoders,  making it quite a challenge to establish reasoning associations between features and quality}.

To address the aforementioned issues, in this work, we propose \textbf{a LLM-based AIGV visual quality evaluation model called AIGVEval}, which consists of three encoders and a quality regression module incorporating the LLM. The overall architecture of the model is illustrated in Figure \ref{fig:fig2}. Specifically, we characterize the visual quality of AIGVs from the following three dimensions: 1) \textbf{video semantics}, focusing on fundamental visual features such as color, texture and scene descriptions. 2) \textbf{technical quality}, similar to the evaluation of non-AIGVs, focuses on identifying spatio-temporal distortions in the video, such as blurriness or shakiness. 3) \textbf{motion quality}, examining the dynamic degree of the video and whether there are content jumps between frames. We design dedicated encoder for each dimension and ultimately employ a LLM-based quality regression module to predict scores. To fully leverage the potential of LLMs in quality assessment tasks, we propose a multi-modal prompt engineering framework, with semantic anchors serving as a pivotal component. This framework effectively aligns the features extracted by the three encoders with the inference space of the LLM, while the semantic anchors facilitate the establishment of associative reasoning across the three feature dimensions. During the training phase, we integrate LoRA fine-tuning techniques~\cite{hu2022lora} to inject task-specific knowledge into the LLM, enabling it to adapt more effectively to AIGV visual quality assessment.

In summary, our contributions can be summarized as follows:

\begin{itemize}
    \item We proposed an AIGV visual quality evaluation model called AIGVEval, which consists of three encoders and a LLM-based quality regression module. The encoders characterize the video from three dimensions: video semantics, technical quality, and motion quality.
    \item To fully leverage the powerful reasoning capabilities of LLMs, we designed a prompt engineering framework specifically tailored for AIGV evaluation tasks. Additionally, we incorporated LoRA fine-tuning during the training phase to further enhance the model's adaptability to AIGV evaluation tasks.
    \item Our method achieved impressive performance in the AIGV visual quality assessment task, securing \textbf{second place} in the NTIRE 2025 Quality Assessment of AI-Generated Content Challenge: Track 2 AI Generated video~\cite{liu2025ntire}.
\end{itemize}
\section{Related Works}
\label{sec:formatting}

\subsection{AIGVs Quality Assessment}

With the rapid advancement of video generation models~\cite{brooks2024video, peng2025open, kong2024hunyuanvideo, singer2022make, hong2022cogvideo, ma2025step, zheng2024open}, researchers have proposed a series of no-reference quality assessment methods specifically designed for evaluating AI-Generated videos. AIGC-VQA~\cite{lu2024aigc} presents a three-branch framework for modeling technical quality, aesthetic quality, and text-video alignment. It leverages ResNet-50~\cite{he2016deep}, ConvNeXt-3D~\cite{liu2022convnet}, and BLIP~\cite{li2022blip} with adapters, and fuses their outputs through a MLP for final score regression. EvalCrafter~\cite{liu2024evalcrafter} introduces a large-scale benchmark with 17 sub-dimensions, including Count Score and Flow Score, using VideoMAE V2~\cite{wang2023videomae}, SDXL~\cite{podell2023sdxl}, and BLIP~\cite{li2022blip} for feature extraction and linear regression. VBench++~\cite{huang2024vbench++} expands evaluation to dimensions such as object consistency, motion smoothness, text-video alignment, and trustworthiness, using UMT~\cite{li2023unmasked}, GRiT~\cite{wu2024grit}, and CLIP~\cite{radford2021learning}.

Temporal coherence represents a critical dimension in evaluating the perceptual quality of AI-generated videos. T2VBench~\cite{ji2024t2vbench} evaluates temporal coherence from event, visual, and narrative perspectives with metrics such as VQAScore~\cite{lin2024evaluating} and ImageReward~\cite{xu2023imagereward}. ChronoMagic-Bench~\cite{yuan2024chronomagic} focuses on time-lapse video consistency and proposes MTScore and CHScore based on object motion and saliency change. GAIA~\cite{chen2024gaia} benchmarks human-object interaction quality using SlowFast network~\cite{feichtenhofer2019slowfast} and temporal attention, generating Flow Score and Action Score. Overall, these methods typically rely on pretrained encoders and structured features combined with lightweight regressors. While performance has improved, their limited reasoning capacity and restricted semantic capacity have motivated the adoption of LLM-based approaches.

\subsection{LLM-Based Video Quality Assessment}

The growing capabilities of large language models~\cite{ji2024t2vbench, yang2024qwen2, achiam2023gpt, team2024gemini, guo2025deepseek} in language modeling, instruction following, and logical reasoning have introduced new paradigms for video quality assessment.
For the quality assessment of non-AIGVs, Q-ALIGN~\cite{wu2023q} converts discrete quality levels into natural language prompts (e.g., “good,” “fair,” “bad”) and employs BLIP2~\cite{li2023blip} to extract quality-aware embeddings for downstream scoring tasks. Q-Instruct~\cite{wu2024q} fine-tunes GPT on feedback-based instructions to improve recognition of low-level distortions. LMM-VQA~\cite{ge2024lmm} performs end-to-end score regression by feeding visual features into a Llama-3-8b-Instruct decoder~\cite{grattafiori2024llama}. Duan et al. proposed Fine-VQ~\cite{duan2024finevq}, which extracts spatial and temporal features of video content using InternViT and SlowFast, while utilizing InternVL2-8B for quality prediction. Additionally, low-rank matrices are added to each layer of InternViT and the LLM to optimize training efficiency. VQA$^2$~\cite{jia2024vqa} utilizes SigLip~\cite{zhai2023sigmoid} and SlowFast~\cite{feichtenhofer2019slowfast} networks to extract spatio-temporal features from videos, employs Qwen-2~\cite{yang2024qwen2} for score inference, and designs a three-stage training pipeline to unify video quality scoring and quality understanding. Wen et al.~\cite{wen2025ensemble} proposed a weighted fusion of the prediction results from four mainstream learning-based BVQA models and PaliGemma~\cite{beyer2024paligemma}. The fused model demonstrated significantly superior performance compared to using MLLM or BVQA models individually on two short-form video datasets.

For the quality assessment of AI-generated videos, VIDEOSCORE~\cite{he2024videoscore} builds upon the Mantis-Idefics2-8B model~\cite{jiang2024mantis}, capable of providing textual descriptions of video quality as well as predicting scores. T2VQA~\cite{kou2024subjective} fuses 3D Swin-T and BLIP features, then applies an LLM to jointly assess alignment and video fidelity. T2VEval~\cite{qi2025comprehensive} employs 3D Swin-T, ConvNeXt 3D, and BLIP to encode technical quality, realness, and text-video consistency respectively, and predicts five-class scores with Vicuna v1.5~\cite{zheng2023judging} along with language explanations. Wang et al. proposed the AIGV-Assessor~\cite{wang2024aigv}, which extracts spatial and temporal features of video content using InternViT and SlowFast, respectively. It employs InternVL2-8B to map visual features into the language space, generating video quality level descriptions and predicting precise quality scores.

In summary, LLM-based methods show strong potential in perceptual modeling, semantic reasoning, and multi-modal alignment, and are becoming central to the evaluation of AI-generated video.

\section{Proposed Method}

\begin{figure*}[hbtp]
    \centering
    \includegraphics[width=0.93\linewidth]{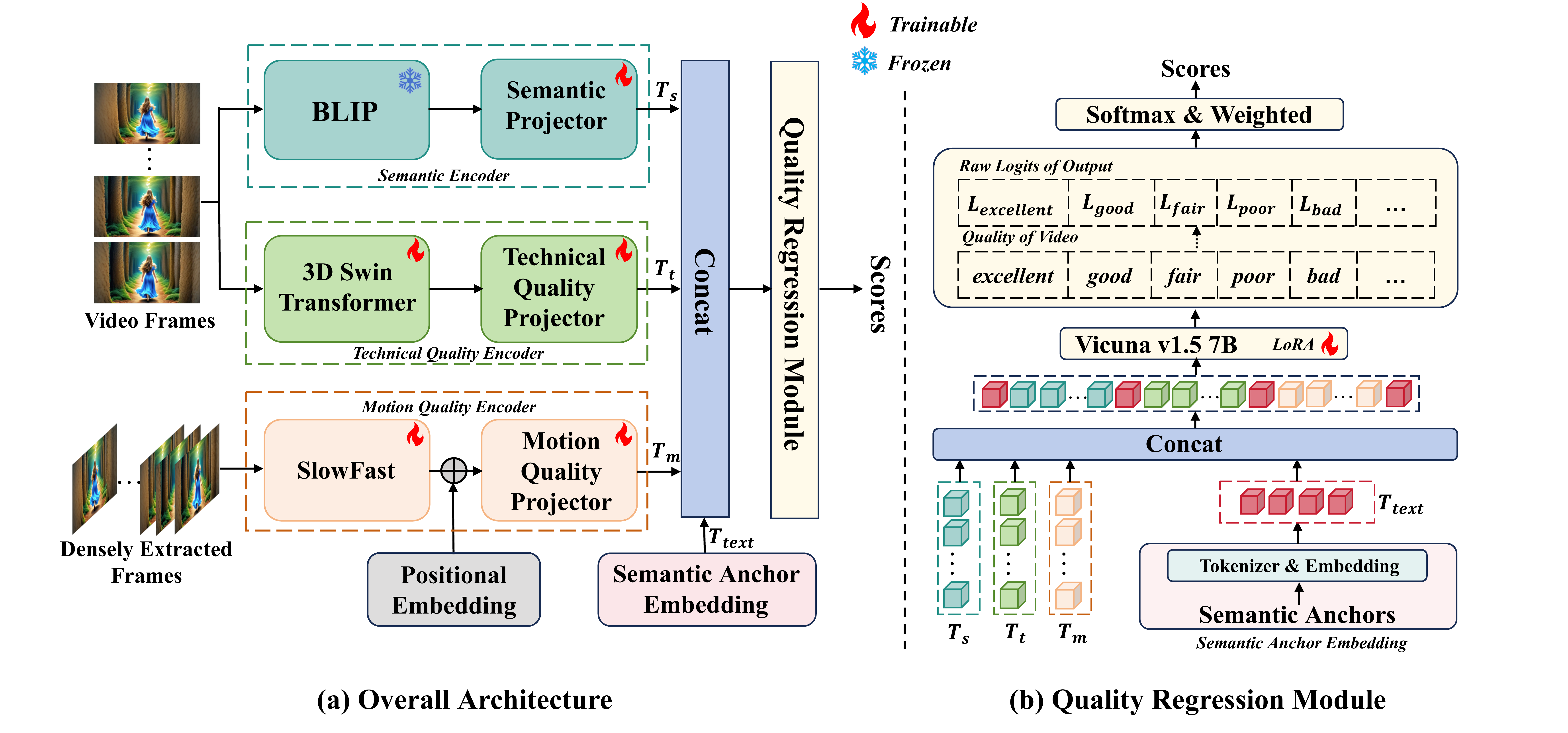}
    \caption{(a) The overall architecture of our proposed AIGVEval, which includes three encoders and an LLM-based quality regression module. (b) A detailed presentation of quality regression module.}
    \label{fig:fig3}
\end{figure*}

The overall framework of AIGVEval is shown in Figure \ref{fig:fig3}, which follows a typical encoder-decoder architecture. For the encoder part, we model AIGV visual quality from three dimensions: video semantics, technical quality, and motion quality. In the decoder part, we employ a 7B version of Vicuna v1.5~\cite{zheng2023judging} for quality regression and design a prompt engineering framework, along with integrating LoRA fine-tuning techniques, to enhance the adaptability of the LLM to the evaluation task. Detailed technical specifications are described below.

\subsection{Semantic Encoder}

Video semantic features typically describe the physical properties of objects in a video, the spatiotemporal relationships between objects, and the content information of the objects. These features belong to high-dimensional characteristics of a video and are closely associated with its low-dimensional features, such as brightness, color, and texture. Moreover, the distortion of semantic features in different video content has varying impacts on human perception of video quality. Therefore, accurate modeling and evaluation of semantic features are critical for video quality analysis.

In this work, we select the visual backbone of BLIP~\cite{li2022blip} as the semantic feature extractor $F_s(\cdot)$. BLIP is a model specifically designed for multi-modal tasks, leveraging joint pre-training of visual and language modalities to achieve outstanding performance in semantic understanding and generation tasks. Its core strength lies in its robust vision-language alignment capabilities, enabling it to extract rich semantic information from images or videos and convert it into high-quality embedding representations.

In the implementation, we first sample frame sequences from the AIGV and input them into BLIP frame by frame. After that, we use a specially designed projection layer $Proj_{s}$ to map features of each frame into the inference space of the LLM, obtaining the semantic tokens of the video, denoted as \( T_s \):

\begin{equation}
    T_s = Proj_{s}(F_s(V)), T_s \in R^{T \times 4096}
\end{equation}

\subsection{Technical Quality Encoder}

Considering the outstanding performance of DOVER~\cite{wu2023exploring} on multiple video quality assessment datasets, we adopt its technical branch backbone network as the technical quality feature extractor $F_t(\cdot)$. Specifically, the extractor utilizes a 3D Swin Transformer network pretrained on Kinetics-400 dataset. For a video \(V\) of a given size $T \times W \times H$, we first perform Grid Mini-patch Sampling (GMS)~\cite{wu2022fast} to avoid detail loss caused by downsampling, thereby obtaining the video segments \(V'\):

\begin{equation}
    V' = GMS(V), V \in R^{T \times W \times W}, V' \in R^{T \times 224 \times 224}
\end{equation}
GMS is an efficient video sampling method designed to maximize the retention of video quality information while reducing computational costs. GMS first divides video frames into uniform grids of size $G_f \times G_f$, ensuring global quality coverage across the frame (In this work, we set $G_f = 32$). Then, a mini-patch of the original resolution is randomly sampled from each grid to preserve local texture information, such as blurriness and noise.

Videos processed by GMS are first passed through $F_t(\cdot)$ to extract quality features, which are then mapped to the LLM's inference space via a quality projection layer $Proj_t$, ultimately generating technical quality tokens \( T_t \):

\begin{equation}
    T_t = Proj_t(F_t(V')), T_t \in R^{T \times 4096}
\end{equation}

The technical quality encoder can effectively extract low-level local features and capture the distortions present in the videos, making it beneficial for evaluating technical quality. 

\subsection{Motion Quality Encoder}

In this paper, we adopt the SlowFast-R50 model~\cite{feichtenhofer2019slowfast} (denoted as \( F_m(\cdot) \)) as the motion quality feature extractor. SlowFast is a dual-branch architecture designed for video understanding tasks. The slow branch processes video spatial information at a lower frame rate, capturing high-resolution static features, while the fast branch processes temporal dynamic information at a higher frame rate, focusing on fine-grained features of fast motion. Through this dual-scale modeling strategy, SlowFast can effectively integrates spatial and temporal information, providing rich and precise representations for modeling motion quality.

Specifically, for a video \( V \in R^{T \times W \times H} \), we first perform dense sampling to obtain densely sampled frames. These frames are then divided into two paths based on different temporal resolutions: the slow path and the fast path. For the slow path, \( T=8 \) frames are uniformly sampled from the densely sampled frames, denoted as \( V_{\text{slow}} \in R^{8 \times W \times H} \), capturing static spatial features. For the fast path, frames are sampled at a higher rate, selecting \( \alpha T \) frames, where \( \alpha \) is set to 4 in this paper, resulting in \( V_{\text{fast}} \in R^{32 \times W \times H} \), which focuses on dynamic temporal features. This division allows the model to effectively capture both spatial and temporal information at different scales. Additionally, positional encoding is applied by performing token-wise addition on the motion tokens and learnable absolute positional embeddings. Finally, a motion projection layer \( Proj_{m} \) is used to align the motion tokens with the inference space of LLMs. 

Based on this, we can obtain the motion quality tokens \( T_m \):

\begin{equation}
    T_m = Proj_{m}(F_m(V_{\text{slow}}, V_{\text{fast}}) + PE), T_m \in R^{\alpha T \times 4096}
\end{equation}

\subsection{Prompt Design}

Prompts serve as a crucial bridge connecting video features with the capabilities of LLMs. The quality of prompt design directly affects the model's ability to understand, analyze, and evaluate video quality. In previous works using LLM prompts for evaluation, the prompts were generally straightforward and simple, such as ``Please rate the quality of this video" in T2VQA. On the other hand, Q-Eval-Score simulated the reasoning stage of the human evaluation process by designing a chain of thought for evaluation in its prompts. However, these prompt designs overlooked the physical meaning of the features fed into LLMs, resulting in the model failing to fully utilize the multi-dimensional feature information provided by the encoder. Instead, the model relied solely on the semantic guidance of the prompts for superficial reasoning.

Inspired by VQA$^2$~\cite{jia2024vqa}, we explicitly embedded the tokens extracted by the encoders into the prompt text to convey the specific feature information corresponding to each evaluation dimension to the LLM. This allows the model not only to understand the semantic meaning of the evaluation dimensions but also to clearly identify the available feature content and its physical significance. Specifically, we designed a series of \textbf{semantic anchors} in the prompts to explicitly specify the content of tokens for each dimension. We defined four semantic anchors to indicate tokens of each dimension and descript the task, including:
\begin{itemize}
    \item \textbf{Video Semantics anchor}: ``The key frames of this video are".
    \item \textbf{Technical quality anchor}: ``the technical quality features of the video are".
    \item \textbf{Motion quality anchor}: ``the motion quality features of the video are".
    \item \textbf{Task description}: ``Please assess the quality of this video".
\end{itemize}

The final designed prompt is as follows:  \emph{``The key frames of this video are:'' + \textbf{Semantic TOKEN} + ``, the technical quality features of the video are:" + \textbf{Technical Quality TOKEN} + ``and the motion quality features of the video are:" + \textbf{Motion Quality TOKEN} + ``. Please assess the quality of this video"}. Through this prompt design, the model can better understand the semantics of multi-dimensional features and leverage the information provided by the encoder for deeper reasoning, thereby enhancing its ability to assess the quality of AIGVs.

\subsection{Semantic Anchor Embedding}

We tokenized and encoded the semantic anchors to generate the textual representation $T_{text}$. By utilizing the relative positional relationships, the visual tokens and textual tokens are concatenated to obtain the final multi-modal representation. The resulting token not only delivers task instructions to the LLM but also offers explicit explanations for the tokens in each dimension, effectively guiding the model in predicting AIGV visual quality.

\subsection{Quality Regression Module}

The output of LLM is the probability distribution of the five quality levels (bad, poor, fair, good, and excellent). We assigned weights of 1-5 to these five quality levels, respectively. Finally, the predicted score was obtained by calculating the softmax of each token and multiplying it by the corresponding weight.
\begin{equation}
s_{pred}=\sum_{i=1}^5i\times softmax(\lambda_i)=\sum_{i=1}^5i\times\frac{e^{\lambda_i}}{\sum_{j=1}^5e^{\lambda_j}}
\end{equation}
where $\lambda_i$ is the probability distribution of the i-th level token.

\section{Experiments}

\subsection{Dataset}

Our experiments were conducted using the Q-Eval dataset provided in the NTIRE 2025 Quality Assessment of AI-Generated Content Challenge: Track 2 AI Generated video. This dataset is divided into three parts: a training set, a validation set, and a test set. The training set includes 23,820 videos along with their corresponding Mean Opinion Scores (MOSs) and textual prompts. The validation and test datasets contain 3,402 and 6,807 videos, respectively, and their corresponding textual prompts.

It is important to note that, according to the description of Q-Eval, the MOSs provided in the competition dataset are solely based on the visual quality of the AIGVs and do not involve an assessment of text-video consistency.

\subsection{Implement Details}

For the senmatic encoder, we used the BLIP's vision encoder~\cite{li2022blip} to extract features, while the technical quality extractor was initialized with weights pre-trained by 3D Swin-T~\cite{liu2022video} on the Kinetics-400 dataset~\cite{kay2017kinetics}. To capture motion quality feature, we employed SlowFast R50~\cite{feichtenhofer2019slowfast} for feature extraction. The LLM used in the quality regression module is the 7B version of Vicuna v1.5~\cite{zheng2023judging}, which is fine-tuned from LLaMA2. 

During the training phase, we implemented the code using PyTorch and employed the Adam optimizer with an initial learning rate of 1e-5 and a decay rate of 0.05. The learning rate was dynamically adjusted using LambdaLR scheduler. The model was trained for 10 epochs on an NVIDIA A800 GPU, including 2.5 warmup epochs. We froze the gradients of the BLIP component while fine-tuning the LLM using Low-Rank Adaptation (LoRA) with a matrix rank of 8, specifically applied to the Key and Query matrices, consistent with the parameters recommended in \cite{hu2022lora}.

Following~\cite{kou2024subjective,qi2025t2vevalbenchmarkdatasetobjective}, we use differentiable Pearson's Linear Correlation Coefficient (PLCC) and rank loss as loss functions. PLCC is a common criterion used for evaluating the correlation between sequences, while rank loss is introduced to help the model distinguish the relative quality of videos better.
\begin{equation}
    L_{plcc} = \frac{1}{2} \left( 1 - \frac{\sum_{i=1}^m (s_i - \bar{s})(y_i - \bar{y})}{\sqrt{\sum_{i=1}^m (s_i - \bar{s})^2} \sqrt{\sum_{i=1}^m (y_i - \bar{y})^2}} \right)
\end{equation}
    
\begin{equation}
    L_{rank} = \frac{1}{m^2} \sum_{i=1}^m \sum_{j=1}^m \left( \max\left( 0, |y_i - y_j| - e(y_i, y_j) \cdot (s_i - s_j) \right) \right)
\end{equation}
    
Here, $y$ and $\bar{y}$ denote the MOSs and their mean value, respectively, while $s$ and $\bar{s}$ represent the prediction scores and corresponding mean value. 
    
The final loss function is defined as:
\begin{equation}
    L = L_{plcc} + \lambda \cdot L_{rank}
\end{equation}
$\lambda$ is a hyperparameter used to balance the different loss functions, which we set to 0.3.

\subsection{Evaluation Metrics}

Similar to traditional No-Reference Video Quality Assessment (NR-VQA), the MOSs are used as the ground truth. The evaluation metrics consist of a combination of PLCC (Pearson’s Linear Correlation Coefficient) and SROCC (Spearman’s Rank Order Correlation Coefficient). PLCC measures the linear correlation between the predicted scores and MOSs, with values ranging from -1 to 1. A value of 1 indicates a perfect positive linear correlation, 0 indicates no linear correlation, and -1 indicates a perfect negative linear correlation. The formula is provided as follows:

\begin{equation}
    PLCC = \frac{\sum_{i=1}^{n}(x_i - \bar{x})(y_i - \bar{y})}{\sqrt{\sum_{i=1}^{n}(x_i - \bar{x})^2 \sum_{i=1}^{n}(y_i - \bar{y})^2}}
\end{equation}
where $x_i$ and $y_i$ represent the predicted scores and the MOSs for the $i$-th sample, respectively, and $\bar{x}$ and $\bar{y}$ are the mean values of the predicted scores and the MOSs, respectively.

SROCC, on the other hand, measures the rank correlation between the predicted scores and the MOSs. It is calculated by comparing the ranks of the predicted scores and the MOSs for each sample. The formula is given by:
\begin{equation}
    SROCC = 1 - \frac{6 \sum_{i=1}^{n} d_i^2}{n(n^2 - 1)}
\end{equation}
where $d_i$ represents the difference in ranks between the predicted scores and the MOSs for the $i$-th sample, and $n$ is the total number of samples. A value of 1 indicates a perfect rank correlation, 0 indicates no rank correlation, and -1 indicates a perfect negative rank correlation.

The overall score, \emph{MainScore}, is obtained by ignoring the sign and reporting the average of absolute values $(PLCC + SROCC)/2$.

\subsection{Experimental Results}

\begin{table}[htbp]
    \centering
    \caption{Leaderboard of NTIRE 2025 Quality Assessment of AI-Generated Content Challenge: Track 2 AI Generated video.}
    \label{tab:leaderboard}
    \scalebox{0.9}{ 
        \begin{tabular}{c|c|c|c|c}
            \hline
            \textbf{Rank} & \textbf{Team} & \textbf{MainScore $\uparrow$} & \textbf{PLCC $\uparrow$} & \textbf{SROCC $\uparrow$} \\
            \hline
            1 & SLCV & 0.665 & 0.667 & 0.662 \\
            2 & \textbf{CUC-IMC} & \textbf{0.631} & \textbf{0.654} & \textbf{0.608} \\
            3 & opdai & 0.590 & 0.595 & 0.585 \\   
            4 & Magnolia & 0.589 & 0.584 & 0.593\\
            5 & AIGC VQA & 0.561 & 0.573 & 0.549 \\
            6 & SJTU-MOE-AI & 0.547 & 0.540 & 0.553 \\
            \hline
        \end{tabular}
    }
\end{table}

NTIRE 2025 Quality Assessment of AI-Generated Content Challenge: Track 2 AI Generated video aimed to foster the development of effective methodologies for assessing the quality of AI-generated videos. Our method achieved \textbf{second place} in this track, as shown in Table \ref{tab:leaderboard}.

We conducted comparative experiments using the specified test dataset to evaluate our proposed method against the following approaches: DOVER~\cite{wu2023exploring}, SimpleVQA~\cite{sun2022deep}, FAST-VQA~\cite{wu2022fast}, Q-Align~\cite{wu2023q}, T2VQA~\cite{kou2024subjective}, Q-Eval-Score~\cite{zhang2025q}, and T2VEval~\cite{qi2025t2vevalbenchmarkdatasetobjective}. Among them, DOVER, SimpleVQA, FAST-VQA, and Q-Align are currently advanced and representative VQA methods, while T2VQA, Q-Eval-Score, and T2VEval are specifically designed for AIGV quality evaluation. To ensure fair and consistent evaluation, all models were trained on the same training dataset and tested on the competition platform. The results in Table \ref{tab:comparison} clearly demonstrate that our proposed method achieves superior performance compared to the existing state-of-the-art methods.

\begin{table}[htbp]
    \centering
    \caption{Comparison of Different Methods on the Test Dataset. \textbf{\textcolor{red}{Red}}: the best}
    \label{tab:comparison}
    \scalebox{0.9}{
        \begin{tabular}{c|c|c|c}
            \hline
            \textbf{Method} & \textbf{MainScore $\uparrow$} & \textbf{PLCC $\uparrow$} & \textbf{SROCC $\uparrow$} \\
            \hline
            DOVER~\cite{wu2023exploring} & 0.506 & 0.505 & 0.506 \\
            SimpleVQA~\cite{sun2022deep} & 0.365 & 0.346 & 0.385 \\
            FAST-VQA~\cite{wu2022fast} & 0.108 & 0.104 & 0.111 \\
            Q-Align~\cite{wu2023q} & 0.438 & 0.416 & 0.460 \\
            T2VQA~\cite{kou2024subjective} & 0.516 & 0.516 & 0.516 \\
            Q-Eval-Score~\cite{zhang2025q} & 0.474 & 0.486 & 0.464\\
            T2VEval~\cite{qi2025t2vevalbenchmarkdatasetobjective} & 0.525 & 0.516 & 0.535 \\
            \hline
            \textcolor{red}{\textbf{AIGVEval}} & \textcolor{red}{\textbf{0.631}} & \textcolor{red}{\textbf{0.654}} & \textcolor{red}{\textbf{0.608}} \\
            \hline
        \end{tabular}
    }
\end{table}

\begin{table*}[htbp]
    \centering
    \caption{Performance comparison of ablation studies. \textbf{\textcolor{red}{Red}}: the best}
    \label{tab:ablation}
    \begin{tabular}{ccccc}
        \toprule
        \textbf{Categories of ablation experiments} & \textbf{Models} & \textbf{MainScore $\uparrow$} & \textbf{PLCC $\uparrow$} & \textbf{SROCC $\uparrow$}  \\
        \midrule
        \multirow{3}{*}{Ablation of Multi-Dimensional Encoder} 
        & Without Tec Quality & 0.546 & 0.550 & 0.541 \\
        & Without Motion Quality & 0.610 & 0.615 & 0.606 \\
        & Without Video Semantics & 0.655 & 0.666 & 0.645 \\
        \midrule
        \multirow{2}{*}{Ablation of Semantic Anchors} 
        & Directly Concat & 0.636 & 0.650 & 0.621 \\
        & Fusion & 0.667 & 0.679 & 0.656 \\
        \midrule
        \multirow{1}{*}{\textcolor{red}{\textbf{Our Method}}}
        & \textcolor{red}{\textbf{AIGVEval}} & \textcolor{red}{\textbf{0.695}} & \textcolor{red}{\textbf{0.706}} & \textcolor{red}{\textbf{0.684}} \\
        \bottomrule
    \end{tabular}
\end{table*}

\begin{table}[htbp]
    \centering
    \caption{Performance Comparison on T2VQA-DB\cite{kou2024subjective}. \textbf{\textcolor{red}{Red}}: the best, \textbf{Bold}: ours}
    \label{tab:otherdataset}
    \scalebox{0.85}{ 
        \begin{tabular}{c|c|c|c|c}
            \hline
            \textbf{Model} & \textbf{PLCC $\uparrow$} & \textbf{SROCC $\uparrow$} & \textbf{KRCC $\uparrow$} & \textbf{RMSE $\downarrow$} \\
            \hline
            CLIPSim~\cite{wu2021godiva} & 0.1277 & 0.1047 & 0.0702 & 21.683 \\
            BLIP~\cite{li2022blip} & 0.1860 & 0.1659 & 0.1112 & 18.373 \\
            ImageReward~\cite{xu2024imagereward} & 0.2121 & 0.1875 & 0.1266 & 18.243 \\   
            ViCLIP~\cite{wang2023internvid} & 0.1449 & 0.1162 & 0.0781 & 21.655\\ 
            UMTScore~\cite{liu2024fetv} & 0.0721 & 0.0676 & 0.0453 & 22.559\\
            SimpleVQA~\cite{sun2022deep} & 0.6338 & 0.6275 & 0.4466 & 11.163 \\
            BVQA~\cite{li2022blindly} & 0.7486 & 0.7390 & 0.5487 & 15.645 \\
            FAST-VQA~\cite{wu2022fast} & 0.7295 & 0.7173 & 0.5303 & 10.595\\
            DOVER~\cite{wu2023exploring} & 0.7693 & 0.7609 & 0.5704 & 9.8072 \\
            T2VQA~\cite{kou2024subjective} & 0.8066 & 0.7965 & 0.6058 & 9.0221\\
            \textbf{\textcolor{red}{T2VEval~\cite{qi2025t2vevalbenchmarkdatasetobjective}}} &\textbf{\textcolor{red}{0.8175}} & \textbf{\textcolor{red}{0.8049}} & \textbf{\textcolor{red}{0.6159}} & \textbf{\textcolor{red}{8.6133}} \\
            \textbf{AIGVEval} & \textbf{0.7494} & \textbf{0.7636} & \textbf{0.5608 }& \textbf{10.034 }\\
            \hline
        \end{tabular}
    }
\end{table}

To further verify the effectiveness of our method, we conducted further tests on the open source dataset T2VQA-DB~\cite{kou2024subjective}. The experimental results are shown in the Table \ref{tab:otherdataset}. It can be seen that AIGVEval also shows competitive performance on T2VQA-DB, but there is still a gap compared to the SOTA method (like T2VQA and T2VEval). This may be because T2VQA-DB considers the consistency of video and textual prompts, which is not considered in our method.

\subsection{Ablation Study}

To validate the effectiveness of key components in our framework, we conducted ablation studies on the validation set. The ablation studies aim to investigate the contributions of the Technical Quality Encoder, Motion Quality Encoder, Semantic Encoder, and the effect of semantic anchors. Main results are summarized in Tabel \ref{tab:ablation}.

\textbf{Ablation Experiments of Technical Quality Encoder.} The technical quality encoder extracts low-level distortion features critical for evaluating the video quality in temporal and spatial domains. When removing this encoder, the MainScore drops from 0.695 to 0.546 (see Tabel \ref{tab:ablation}, Line 1 vs. Line 6). This demonstrates that the distortions in technical quality significantly impact the overall visual perceptual quality, and our 3D-Swin Transformer-based encoder effectively captures these features.

\textbf{Ablation Experiments of Motion Quality Encoder.} The motion quality encoder models dynamic degree using SlowFast-R50, focusing on motion fluency and temporal consistency. Ablating this component leads to a performance decrease from 0.695 to 0.610 (Tabel \ref{tab:ablation}, Line 2 vs. Line 6). The fast branch’s high temporal resolution proves essential for detecting jitter and unstable motions in AIGVs.

\textbf{Ablation Experiments of Semantic Encoder.} The BLIP-based semantic encoder focuses on extracting fundamental visual features from the video, such as color, texture, and structure. Removing semantic features causes a notable performance decrease from 0.695 to 0.655 (Table \ref{tab:ablation}, Line 3 vs. Line 6). Although the impact of ablating the semantic encoder is relatively smaller compared to removing the technical quality or motion quality encoders, this result still highlights the beneficial role semantic information plays in assisting the LLM to effectively evaluate the quality of AIGVs.

\textbf{Ablation Experiments of Semantic Anchors.} To verify the effectiveness of semantic anchors, we conducted experiments comparing our approach against two baseline strategies: (1) directly concatenating the outputs from the Semantic Encoder, Technical Quality Encoder, and Motion Quality Encoder, followed by regression scoring with the LLM; and (2) using cross-attention fusion to combine the outputs from the three encoders before feeding them into the LLM. The MainScore decreased significantly from 0.695 to 0.636 and 0.667, respectively (see Tabel \ref{tab:ablation}, Line 4, Line 5 vs. Line 6). These results clearly indicate that employing semantic anchors effectively enables the LLM to understand the underlying meanings of the three encoder outputs, thus substantially enhancing the evaluation performance.
\section{Conclusion}

In summary, we proposed AIGVEval, a LLM-based multi-dimensional objective evaluation model for AIGVs. Our approach models AIGV visual quality across three key dimensions: technical quality, motion quality, and video semantics, with specifically designed encoders tailored to each dimension. Experimental results validate the effectiveness and rationale behind these three proposed dimensions. Building on this, we introduced semantic anchors to explicitly convey the tokens of each dimension to the LLM, enabling more effective associative reasoning between features and quality. During the training phase, we leveraged LoRA fine-tuning to infuse AIGV visual evaluation knowledge into the LLM, enabling it to better adapt to AIGV quality assessment tasks. Our method secured second place in the NTIRE 2025 Quality Assessment of AI-Generated Content Challenge: Track 2 AI Generated video. The effectiveness of our approach is further supported by comparative and ablation studies.

\section{Acknowledgments}
This work is supported by \textbf{Public Computing Cloud, CUC}.

{
    \small
    \bibliographystyle{ieeenat_fullname}
    \bibliography{main}
}

\end{document}